\documentclass[runningheads]{llncs}
\usepackage{microtype}
\usepackage{graphicx}
\usepackage{subfigure}
\usepackage{booktabs} 
\usepackage{todonotes}
\usepackage{amsmath}
\usepackage{amssymb}
\usepackage{tikz}
\usepackage{csquotes}
\usepackage{float}
\usepackage{bbm}
\usepackage{dsfont}
\usepackage{hyperref}
\newcommand\MyBox[1]{%
    \fbox{\parbox[c][1.5cm][c]{2.5cm}{ #1}}
}
\newcommand\MyVBox[1]{%
    \parbox[c][1cm][c]{1cm}{\centering\bfseries #1}%
}  
\newcommand\MyHBox[2][\dimexpr3cm+2\fboxsep\relax]{%
    \parbox[c][1cm][c]{#1}{\centering\bfseries #2}%
}  
\newcommand\MyTBox[4]{%
    \MyVBox{#1}
    \MyBox{#2}\hspace*{-\fboxrule}%
    \MyBox{#3}\par\vspace{-\fboxrule}%
}  
\newcommand*\rot{\rotatebox{90}}

\begin{document}
\title{General Pitfalls of Model-Agnostic Interpretation Methods for Machine Learning Models\thanks{
This work is funded by the Bavarian State Ministry of Science and the Arts (coordinated by the Bavarian Research Institute for Digital Transformation (bidt)), by the German Federal Ministry of Education and Research (BMBF) under Grant No. 01IS18036A,  by the German Research Foundation (DFG), Emmy Noether Grant 437611051, and by the Graduate School of Systemic Neurosciences (GSN) Munich.
The authors of this work take full responsibilities for its content.
}}
\titlerunning{General Pitfalls of Model-Agnostic Interpretation}
%
\author{Christoph Molnar\inst{1,7}\orcidID{0000-0003-2331-868X} \and
Gunnar König\inst{1,4}\orcidID{0000-0001-6141-4942} \and
Julia Herbinger\inst{1}\orcidID{0000-0003-0430-8523} \and
Timo Freiesleben\inst{2,3}\orcidID{0000-0003-1338-3293} \and
Susanne Dandl\inst{1}\orcidID{0000-0003-4324-4163} \and
Christian A. Scholbeck\inst{1}\orcidID{0000-0001-6607-4895} \and
Giuseppe Casalicchio\inst{1}\orcidID{0000-0001-5324-5966} \and
Moritz Grosse-Wentrup\inst{4,5,6}\orcidID{0000-0001-9787-2291} \and
Bernd Bischl\inst{1}\orcidID{0000-0001-6002-6980} }
\authorrunning{C. Molnar et al.}
%
\institute{Department of Statistics, LMU Munich, Munich, Germany \and
Munich Center for Mathematical Philosophy, LMU Munich, Munich, Germany \and
Graduate School of Systemic Neurosciences, LMU Munich, Munich, Germany \and
Research Group Neuroinformatics, Faculty for Computer Science, University of Vienna, Vienna, Austria \and
Research Platform Data Science @ Uni Vienna, Vienna, Austria \and
Vienna Cognitive Science Hub, Vienna, Austria \and
Leibniz Institute for Prevention Research and Epidemiology – BIPS GmbH, Bremen, Germany 
\\
\email{\{christoph.molnar.ai\}@gmail.com}}
\maketitle              
\begin{abstract}
  An increasing number of model-agnostic interpretation techniques for machine learning (ML) models such as partial dependence plots (PDP), permutation feature importance (PFI) and Shapley values provide insightful model interpretations, but can lead to wrong conclusions if applied incorrectly.
  We highlight many general pitfalls of ML model interpretation, such as using interpretation techniques in the wrong context, interpreting models that do not generalize well, ignoring feature dependencies, interactions, uncertainty estimates and issues in high-dimensional settings, or making unjustified causal interpretations, and illustrate them with examples.
  We focus on pitfalls for global methods that describe the average model behavior, but many pitfalls also apply to local methods that explain individual predictions.
  Our paper addresses ML practitioners by raising awareness of pitfalls and identifying solutions for correct model interpretation, but also addresses ML researchers by discussing open issues for further research.
\end{abstract}

\section{Introduction}
\label{sec:introduction}
In recent years, both industry and academia have increasingly shifted away from parametric models, such as generalized linear models, and towards non-parametric and non-linear machine learning (ML) models such as random forests, gradient boosting, or neural networks. The major driving force behind this development has been a considerable outperformance of ML over traditional models on many prediction tasks \cite{fernandez2014we}. In part, this is because most ML models handle interactions and non-linear effects automatically. While classical statistical models -- such as generalized additive models (GAMs) -- also support the inclusion of interactions and non-linear effects, they come with the increased cost of having to (manually) specify and evaluate these modeling options.
The benefits of many ML models are partly offset by their lack of interpretability, which is of major importance in many applications. 
For certain model classes (e.g. linear models), feature effects or importance scores can be directly inferred from the learned parameters and the model structure.
In contrast, it is more difficult to extract such information from complex non-linear ML models that, for instance, do not have intelligible parameters and are hence often considered black boxes.
However, model-agnostic interpretation methods allow us to harness the predictive power of ML models while gaining insights into the black-box model.
These interpretation methods are already applied in many different fields.
Applications of interpretable machine learning (IML) include understanding pre-evacuation decision-making \cite{zhao2020modelling} with partial dependence plots \cite{friedman1991multivariate},
inferring behavior from smartphone usage \cite{Stachl2020,stachl2019behavioral} with the help of permutation feature importance \cite{strobl2008conditional} and accumulated local effect plots \cite{apley2016visualizing}, or understanding the relation between critical illness and health records \cite{lauritsen2020explainable} using Shapley additive explanations (SHAP) \cite{lundberg2017unified}.
Given the widespread application of interpretable machine learning, it is crucial to highlight potential pitfalls, that, in the worst case, can produce incorrect conclusions.

This paper focuses on pitfalls for model-agnostic IML methods, i.e. methods that can be applied to any predictive model.
Model-specific methods, in contrast are tied to a certain model class (e.g. saliency maps \cite{kadir_saliency} for gradient-based models, such as neural networks), and are mainly considered out-of-scope for this work. 
We focus on pitfalls for global interpretation methods, which describe the expected behavior of the entire model with respect to the whole data distribution.
However, many of the pitfalls also apply to local explanation methods, which explain individual predictions or classifications.
Global methods include the partial dependence plot (PDP) \cite{friedman1991multivariate}, partial importance (PI) \cite{casalicchio2019visualfi}, accumulated local affects (ALE) \cite{apley2016visualizing}, or the permutation feature importance (PFI) \cite{breiman2001random,fisher2019all,casalicchio2019visualfi}. 
Local methods include the individual conditional expectation (ICE) curves \cite{goldstein2015peeking}, individual conditional importance (ICI) \cite{casalicchio2019visualfi}, local interpretable model-agnostic explanations (LIME) \cite{ribeiro2016should}, Shapley values \cite{vstrumbelj2014explaining} and SHapley Additive exPlanations (SHAP) \cite{lundberg2017unified,lundberg2018consistent} or counterfactual explanations \cite{wachter2017counterfactual,dandl2020multi}.
Furthermore, we distinguish between feature effect and feature importance methods. A feature effect indicates the direction and magnitude of a change in predicted outcome due to changes in feature values. Effect methods include Shapley values, SHAP, LIME, ICE, PDP, or ALE. Feature importance methods quantify the contribution of a feature to the model performance (e.g. via a loss function) or to the variance of the prediction function. Importance methods include the PFI, ICI, PI, or SAGE. See Figure~\ref{fig:iml_quadrants} for a visual summary.\\
\begin{figure}[t]
        \scriptsize
    \begin{center}
    {

        \offinterlineskip

       \raisebox{-3.2cm}[0pt][0pt]{
           \parbox[c][3pt][c]{0cm}{\hspace{-3.5cm}\rot{\textbf{Feature}}\\[10pt]}}\par


    \hspace*{1cm}\MyHBox{Local}\hspace*{-0.3cm}\MyHBox{Global}\par
        \MyTBox{\rot{Effects}}{ICE\\LIME\\Counterfactuals\\Shapley Values\\SHAP}{PDP\\ALE}

        \MyTBox{\rot{Importance}}{ICI}{PI\\PFI\\SAGE}

    }
\end{center}
\caption{Selection of popular model-agnostic interpretation techniques, classified as local or global, and as effect or importance methods.
\label{fig:iml_quadrants}
}
\end{figure}
\normalsize
The interpretation of ML models can have subtle pitfalls, such as when features have dependencies or when non-causal features are used.
Since many of the interpretation methods work by similar principles of manipulating data and \enquote{probing} the model \cite{scholbeck_sipa}, they also share many pitfalls.
ML models usually contain non-linear effects and higher-order interactions. 
Therefore, lower-dimensional or linear approximations can be inappropriate and can result in misleading masking effects.
Furthermore, model-agnostic interpretation methods can \textit{technically} be applied to any ML model and data scenario, although the interpretation can suffer in some cases.
For example, the interpretation method can be applied regardless of the model's performance on test data, but we can only draw solid conclusions about the data when it generalizes well.
Alternatively, a simpler model would suffice for some prediction tasks, and thus ML models plus model-agnostic interpretation techniques add unnecessary complexity to the interpretation in these scenarios.
Model-agnostic methods are a versatile tool in a data scientist's toolbox and can produce a seemingly meaningful description of the model. However, these methods do not indicate whether something is amiss.

%
%
\subsubsection{Contributions:} We uncover and review general pitfalls of model-agnostic interpretation techniques.
Each section describes and illustrates a pitfall, reviews possible solutions for practitioners to circumvent the pitfall, and discusses open issues that require further research. The pitfalls are accompanied by illustrative examples for which the code can be found in this repository: \url{https://github.com/compstat-lmu/code_pitfalls_iml.git}. In addition to reproducing our examples, we invite readers to use this code as a starting point for their own experiments and explorations.
%
%

\subsubsection{Related Work:}
Rudin et al. \cite{rudin2021interpretable} present principles for interpretability and discuss challenges for model interpretation with a focus on inherently interpretable models.
Das et al. \cite{das2020opportunities} survey methods for explainable AI and discuss challenges with a focus on saliency maps for neural networks.
A general warning about using and explaining ML models for high stakes decisions has been brought forward by Rudin \cite{rudin2019stop}, in which the author argues against model-agnostic techniques in favor of inherently interpretable models. 
Krishnan \cite{krishnan_against_interpretability} criticizes the general conceptual foundation of interpretability, but does not dispute the usefulness of available methods.
Likewise, Lipton \cite{lipton2018mythos} criticizes interpretable ML for its lack of causal conclusions, trust, and insights, but the author does not discuss any pitfalls in detail.
Specific pitfalls due to dependent features are discussed by Hooker \cite{hooker_generalized_FANOVA} for PDPs and functional ANOVA as well as by Hooker and Mentch \cite{hooker2019please} for feature importance computations.
Hall \cite{hall2018art} discusses recommendations for the application of particular interpretation methods but does not address general pitfalls.

\section{Assuming One-Fits-All Interpretability}
\label{sec:interpretationgoal}

\subsubsection{Pitfall:} 
Assuming that a single IML method fits in all interpretation contexts can lead to dangerous misinterpretation.
IML methods condense the complexity of ML models into human-intelligible descriptions that only provide insight into specific aspects of the model and data. The vast number of interpretation methods make it difficult for practitioners to choose an interpretation method that can answer their question. Due to the wide range of goals that are pursued under the umbrella term \enquote{interpretability}, the methods differ in which aspects of the model and data they describe. 

%
For example, there are several ways to quantify or rank the features according to their relevance. The relevance measured by PFI can be very different from the relevance measured by the SHAP importance. 
If a practitioner aims to gain insight into the relevance of a feature regarding the model's generalization error, a loss-based method (on unseen test data) such as PFI should be used. If we aim to expose which features the model relies on for its prediction or classification -- irrespective of whether they aid the model's generalization performance -- PFI on test data is misleading.
In such scenarios, one should quantify the relevance of a feature regarding the model's prediction (and not the model's generalization error) using methods like the SHAP importance \cite{lundberg2020local}.

\begin{figure}
    \centering
    \includegraphics[width=\textwidth]{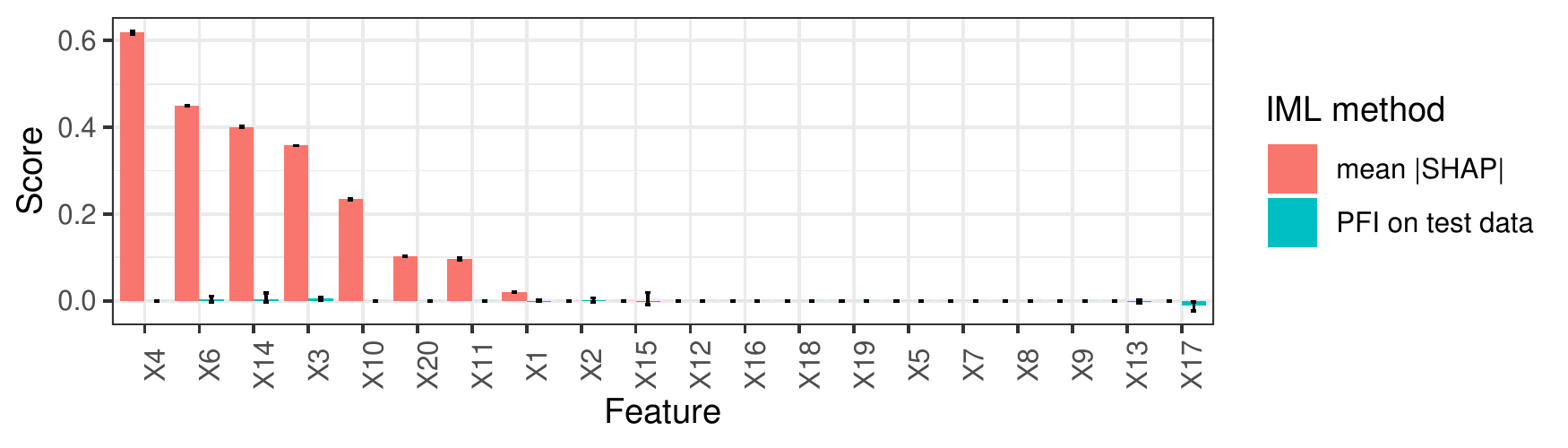}
    \caption{\textbf{Assuming one-fits-all interpretability}. A default \texttt{xgboost} regression model that minimizes the mean squared error (MSE) was fitted on 20 independently and uniformly distributed features to predict another independent, uniformly sampled target. In this setting, predicting the (unconditional) mean $\mathds{E}[Y]$ in a constant model is optimal. The learner overfits due to small training data size. Mean marginal SHAP (red, error bars indicate $0.05$ and $0.95$ quantiles) exposes all mechanistically used features. In contrast, PFI on test data (blue, error bars indicate $0.05$ and $0.95$ quantiles) considers all features to be irrelevant, since no feature contributes to the generalization performance.}
    \label{fig:pfi-test-vs-global-shap}
\end{figure}

We illustrate the difference in Figure \ref{fig:pfi-test-vs-global-shap}. We simulated a data-generating process where the target is completely independent of all features. Hence, the features are just noise and should not contribute to the model's generalization error. Consequently, the features are not considered relevant by PFI on test data. However, the model mechanistically relies on a number of spuriously correlated features. This reliance is exposed by marginal global SHAP importance.

As the example demonstrates, it would be misleading to view the PFI computed on test data or global SHAP as one-fits-all feature importance techniques. Like any IML method, they can only provide insight into certain aspects of model and data.

Many pitfalls in this paper arise from situations where an IML method that was designed for one purpose is applied in an unsuitable context. For example, extrapolation (Section \ref{sec:sampling}) can be problematic when we aim to study how the model behaves under realistic data but simultaneously can be the correct choice if we want to study the sensitivity to a feature outside the data distribution.
\subsubsection{Solution:} The suitability of an IML method cannot be evaluated with respect to one-fits-all interpretability but must be motivated and assessed with respect to well-defined interpretation goals. 
Similarly, practitioners must tailor the choice of the IML method to the interpretation context.
This implies that these goals need to be clearly stated in a detailed manner \textit{before} any analysis -- which is still often not the case. 
\subsubsection{Open Issues:} Since IML methods themselves are subject to interpretation, practitioners must be informed about which conclusions can or cannot be drawn given different choices of IML technique. 
In general, there are three aspects to be considered: (a) an intuitively understandable and plausible algorithmic construction of the IML method to achieve an explanation; (b) a clear mathematical axiomatization of interpretation goals and properties, which are linked by proofs and theoretical considerations to IML methods, and properties of models and data characteristics; (c) a practical translation for practitioners of the axioms from (b) in terms of what an IML method provides and what not, ideally with implementable guidelines and diagnostic checks for violated assumptions to guarantee correct interpretations.
While (a) is nearly always given for any published method, much work remains for (b) and (c).

\section{Bad Model Generalization}
\label{sec:modelfit}

\begin{figure}[t]
    \centering
    \includegraphics[width=\textwidth]{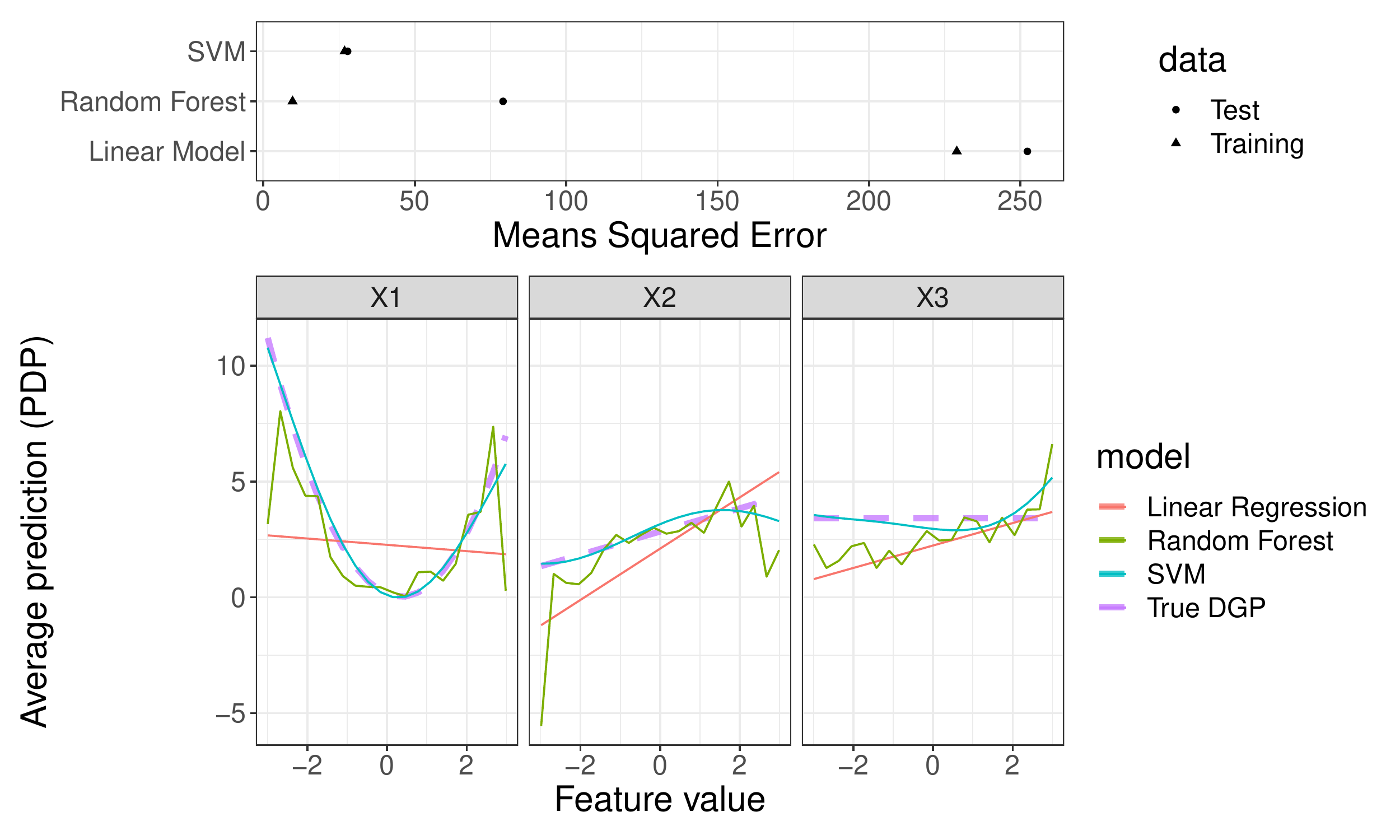}
    \caption{\textbf{Bad model generalization}. \textbf{Top:} Performance estimates on training and test data for a linear regression model (underfitting), a random forest (overfitting) and a support vector machine with radial basis kernel (good fit). The three features are drawn from a uniform distribution, and the target was generated as $Y = X_1^2 + X_2 - 5  X_1  X_2 + \epsilon$, with $\epsilon \sim N(0, 5)$.
    \textbf{Bottom:} PDPs for the data-generating process (DGP) -- which is the ground truth -- and for the three models.}
    \label{fig:generalization}
\end{figure}

\subsubsection{Pitfall:} Under- or overfitting models can result in misleading interpretations with respect to the true feature effects and importance scores, as the model does not match the underlying data-generating process well \cite{good2012common}.
Formally, most IML methods are designed to interpret the model instead of drawing inferences about the data-generating process.
In practice, however, the latter is often the goal of the analysis, and then an interpretation can only be as good as its underlying model. 
If a model approximates the data-generating process well enough, its interpretation should reveal insights into the underlying process.

\subsubsection{Solution:} 
In-sample evaluation (i.e. on training data) should not be used to assess the performance of ML models due to the risk of overfitting on the training data, which will lead to overly optimistic performance estimates.
We must resort to out-of-sample validation based on resampling procedures such as holdout for larger datasets or cross-validation, or even repeated cross-validation for small sample size scenarios.
These resampling procedures are readily available in software \cite{mlr3,scikit-learn}, and well-studied in theory as well as practice \cite{arlot2010cv,bischl2012resampling,simon2007resampling}, although rigorous analysis of cross-validation is still considered an open problem \cite{shalev2014understanding}.
Nested resampling is necessary, when computational model selection and hyperparameter tuning are involved \cite{bischl2021hpo}.
This is important, as the Bayes error for most practical situations is unknown, and we cannot make absolute statements about whether a model already optimally fits the data.

Figure~\ref{fig:generalization} shows the mean squared errors for a simulated example on both training and test data for a support vector machine (SVM), a random forest, and a linear model.
Additionally, PDPs for all models are displayed, which show to what extent each model's effect estimates deviate from the ground truth.
The linear model is unable to represent the non-linear relationship, which is reflected in a high error on both test and training data and the linear PDPs.
In contrast, the random forest has a low training error but a much higher test error, which indicates overfitting.
Also, the PDPs for the random forest display overfitting behavior, as the curves are quite noisy, especially at the lower and upper value ranges of each feature.
The SVM with both low training and test error comes closest to the true PDPs.

\section{Unnecessary Use of Complex Models}
\label{sec:unnecessary}

\subsubsection{Pitfall:} A common mistake is to use an opaque, complex ML model when an interpretable model would have been sufficient, i.e. when the performance of interpretable models is only negligibly worse -- or maybe the same or even better -- than that of the ML model.
Although model-agnostic methods can shed light on the behavior of complex ML models, inherently interpretable models still offer a higher degree of transparency \cite{rudin2019stop} and considering them increases the chance of discovering the true data-generating function \cite{claeskens2008model}. 
What constitutes an interpretable model is highly dependent on the situation and target audience, as even a linear model might be difficult to interpret when many features and interactions are involved.

It is commonly believed that complex ML models always outperform more interpretable models in terms of accuracy and should thus be preferred. However, there are several examples where interpretable models have proven to be serious competitors: More than 15 years ago, Hand \cite{hand2006illusion} demonstrated that simple models often achieve more than 90\% of the predictive power of potentially highly complex models across the UCI benchmark data repository and concluded that such models often should be preferred due to their inherent interpretability;
Makridakis et al. \cite{makridakis2018statistical} systematically compared various ML models (including long-short-term-memory models and multi-layer neural networks) to statistical models (e.g. damped exponential smoothing and the Theta method) in time series forecasting tasks and found that the latter consistently show greater predictive accuracy; Kuhle et al. \cite{kuhle2018comparison} found that random forests, gradient boosting and neural networks did not outperform logistic regression in predicting fetal growth abnormalities; Similarly, Wu et al. \cite{wu2010prediction} have shown that a logistic regression model performs as well as AdaBoost and even better than an SVM in predicting heart disease from electronic health record data;
Baesens et al. \cite{baesens2003benchmarking} showed that simple interpretable classifiers perform competitively for credit scoring, and in an update to the study the authors note that ``the complexity and/or recency of a classifier are misleading indicators of its prediction performance" \cite{lessmann2015benchmarking}.

\subsubsection{Solution:} We recommend starting with simple, interpretable models such as linear regression models and decision trees.
Generalized additive models (GAM) \cite{gam} can serve as a gradual transition between simple linear models and more complex machine learning models.
GAMs have the desirable property that they can additively model smooth, non-linear effects and provide PDPs out-of-the-box, but without the potential pitfall of masking interactions (see Section~\ref{sec:interaction}). The additive model structure of a GAM is specified before fitting the model so that only the pre-specified feature or interaction effects are estimated.
Interactions between features can be added manually or algorithmically (e.g. via a forward greedy search) \cite{caruana2015intelligible}.
GAMs can be fitted with component-wise boosting \cite{schmid2008boosting}. The boosting approach allows to smoothly increase model complexity, from sparse linear models to more complex GAMs with non-linear effects and interactions.
This smooth transition provides insight into the tradeoffs between model simplicity and performance gains.
Furthermore, component-wise boosting has an in-built feature selection mechanism as the model is build incrementally, which is especially useful in high-dimensional settings (see Section~\ref{sec:high-dimensional}).
The predictive performance of models of different complexity should be carefully measured and compared. Complex models should only be favored if the additional performance gain is both significant and relevant -- a judgment call that the practitioner must ultimately make.
Starting with simple models is considered best practice in data science, independent of the question of interpretability \cite{claeskens2008model}.
The comparison of predictive performance between model classes of different complexity can add further insights for interpretation.

%
\subsubsection{Open Issues:} Measures of model complexity allow quantifying the trade-off between complexity and performance and to automatically optimize for multiple objectives beyond performance. Some steps have been made towards quantifying model complexity, such as using functional decomposition and quantifying the complexity of the components \cite{molnar2019quantifying} or measuring the stability of predictions \cite{philipp2018measuring}. However, further research is required, as there is no single perfect definition of interpretability, but rather multiple depending on the context \cite{doshi2017towards,rudin2019stop}.
%
\section{Ignoring Feature Dependence}
\label{sec:dependence}
\subsection{Interpretation with Extrapolation}
\label{sec:sampling}
\subsubsection{Pitfall:}
\begin{figure}[t]
    \centering
    \includegraphics[width=\textwidth]{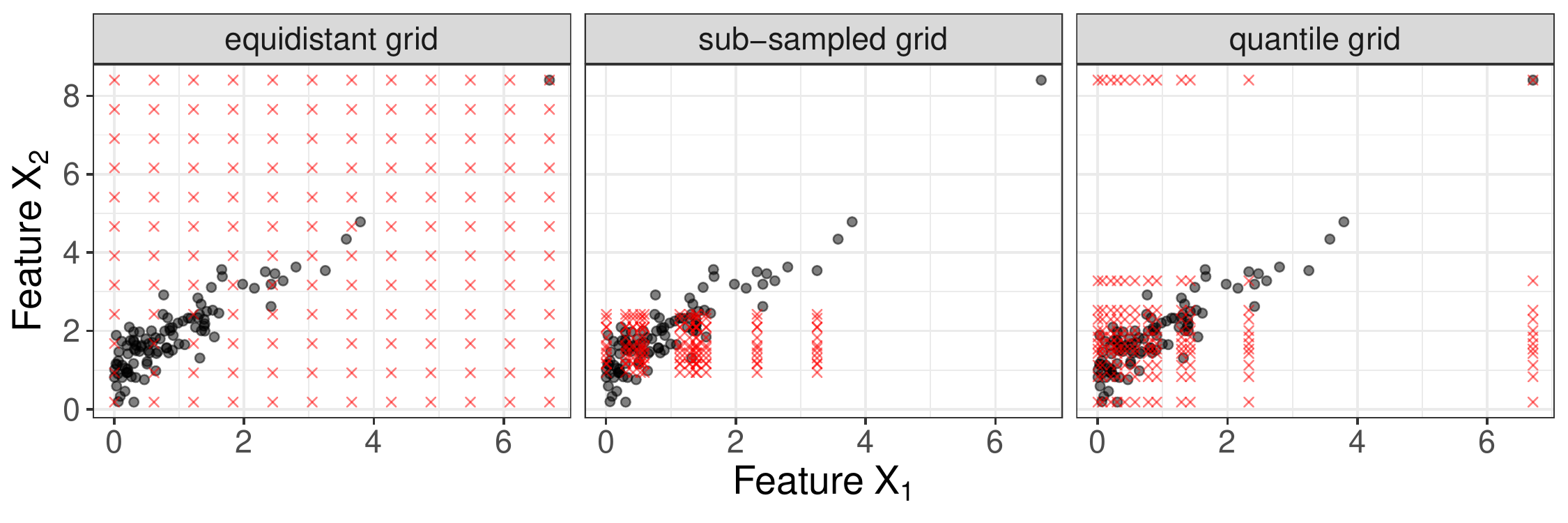}
    \caption{\textbf{Interpretation with extrapolation}. Illustration of artificial data points generated by three different perturbation approaches. The black dots refer to observed data points and the red crosses to the artificial data points.}
    \label{fig:sampling}
\end{figure}

When features are dependent, perturbation-based IML methods such as PFI, PDP, and Shapley values extrapolate in areas where the model was trained with little or no training data, which can cause misleading interpretations \cite{hooker2019please}. This is especially true if the ML model relies on feature interactions \cite{groemping2020model} -- which is often the case.
Perturbations produce artificial data points that are used for model predictions, which in turn are aggregated to produce global interpretations \cite{scholbeck_sipa}.
Feature values can be perturbed by replacing original values with values from an equidistant grid of that feature, with permuted or randomly subsampled values \cite{casalicchio2019visualfi}, or with quantiles. 
We highlight two major issues:
First, if features are dependent, all three perturbation approaches produce unrealistic data points, i.e. the new data points are located outside of the multivariate joint distribution of the data (see Figure~\ref{fig:sampling}).
Second, even if features are independent, using an equidistant grid can produce unrealistic values for the feature of interest.
Consider a feature that follows a skewed distribution with outliers. An equidistant grid would generate many values between outliers and non-outliers.
In contrast to the grid-based approach, the other two approaches maintain the marginal distribution of the feature of interest. 

Both issues can result in misleading interpretations (illustrative examples are given in \cite{hooker2019please,molnar2020model}), since the model is evaluated in areas of the feature space with few or no observed real data points, where model uncertainty can be expected to be very high.
This issue is aggravated if global interpretation methods integrate over such points with the same weight and confidence as for much more realistic samples with high model confidence.
%

\subsubsection{Solution:}
Before applying interpretation methods, practitioners should check for dependencies between features in the data, e.g. via descriptive statistics or measures of dependence (see Section~\ref{sec:correlation}).
When it is unavoidable to include dependent features in the model (which is usually the case in ML scenarios), additional information regarding the strength and shape of the dependence structure should be provided.
Sometimes, alternative interpretation methods can be used as a workaround or to provide additional information. Accumulated local effect plots (ALE) \cite{apley2016visualizing} can be applied when features are dependent, but can produce non-intuitive effect plots for simple linear models with interactions \cite{groemping2020model}.
For other methods such as the PFI, conditional variants exist \cite{candes2018panning,molnar2020model,strobl2008conditional}.
Note, however, that conditional interpretations are often different and should not be used as a substitute for unconditional interpretations (see Section~\ref{sec:conditionalinterpretation}).
Furthermore, dependent features should not be interpreted separately but rather jointly. This can be achieved by visualizing e.g. a 2-dimensional ALE plot of two dependent features, which, admittedly, only works for very low-dimensional combinations. Especially in high-dimensional settings where dependent features can be grouped in a meaningful way, grouped interpretation methods might be more reasonable (see Section~\ref{sec:high-dimensional}).


We recommend using quantiles or randomly subsampled values over equidistant grids.
By default, many implementations of interpretability methods use an equidistant grid to perturb feature values \cite{greenwell2017pdp,molnar2018iml,scikit-learn}, although some also allow using user-defined values.
%

\subsubsection{Open Issues:}
A comprehensive comparison of strategies addressing extrapolation and how they affect an interpretation method is currently missing.
This also includes studying interpretation methods and their conditional variants when they are applied to data with different dependence structures.
\subsection{Confusing Linear Correlation with General Dependence}
\label{sec:correlation}
\subsubsection{Pitfall:} 
\begin{figure}[t]
    \centering
    \includegraphics[width=0.65\textwidth]{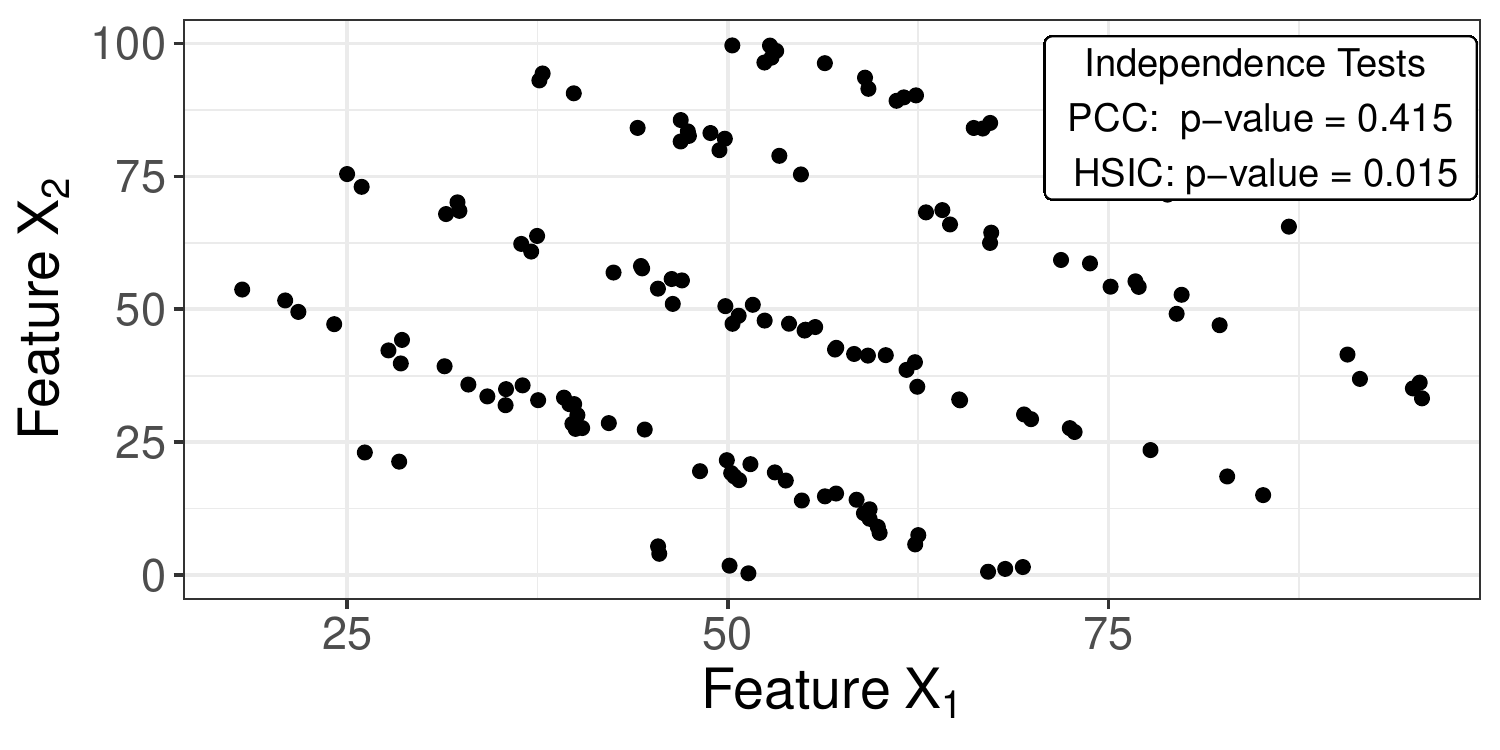}
    \caption{\textbf{Confusing linear correlation with dependence}. Highly dependent features $X_1$ and $X_2$ that have a correlation close to zero. A test ($H_0$: Features are independent) using Pearson correlation is not significant, but for HSIC, the $H_0$-hypothesis gets rejected. Data from \cite{matejka2017same}.}
    \label{fig:association}
\end{figure}

Features with a Pearson correlation coefficient (PCC) close to zero can still be dependent and cause misleading model interpretations (see Figure~\ref{fig:association}).
While independence between two features implies that the PCC is zero, the converse is generally false. The PCC, which is often used to analyze dependence, only tracks linear correlations and has other shortcomings such as sensitivity to outliers \cite{tjstheim2018statistical}.
Any type of dependence between features can have a strong impact on the interpretation of the results of IML methods (see Section~\ref{sec:sampling}).
Thus, knowledge about the (possibly non-linear) dependencies between features is crucial for an informed use of IML methods.


\subsubsection{Solution:}
Low-dimensional data can be visualized to detect dependence (e.g. scatter plots) \cite{matejka2017same}.
For high-dimensional data, several other measures of dependence in addition to PCC can be used.
If dependence is monotonic, Spearman's rank correlation coefficient \cite{liebetrau1983measures} can be a simple, robust alternative to PCC. For categorical or mixed features, separate dependence measures have been proposed, such as Kendall's rank correlation coefficient for ordinal features, or the phi coefficient and Goodman \& Kruskal’s lambda for nominal features \cite{khamis2008measures}.

Studying non-linear dependencies is more difficult since a vast variety of possible associations have to be checked. Nevertheless, several non-linear association measures with sound statistical properties exist.
Kernel-based measures, such as kernel canonical correlation analysis (KCCA) \cite{bach2002kernel} or the Hilbert-Schmidt independence criterion (HSIC) \cite{gretton2005measuring}, are commonly used.
They have a solid theoretical foundation, are computationally feasible, and robust \cite{tjstheim2018statistical}.
In addition, there are information-theoretical measures, such as (conditional) mutual information \cite{cover2012elements} or the maximal information coefficient (MIC) \cite{reshef2011detecting}, that can however be difficult to estimate \cite{walters2009estimation,belghazi2018mutual}. Other important measures are e.g. the distance correlation \cite{szekely2007measuring}, the randomized dependence coefficient (RDC) \cite{lopez2013randomized}, or the alternating conditional expectations (ACE) algorithm \cite{breiman1985estimating}.
In addition to using PCC, we recommend using at least one measure that detects non-linear dependencies (e.g. HSIC).

%
%
%
%
%
\subsection{Misunderstanding Conditional Interpretation}
\label{sec:conditionalinterpretation}
\begin{figure}
\centering

\subfigure[]{
\centering
\begin{tikzpicture}[thick, scale=0.9, every node/.style={scale=0.6, line width=0.3mm, black, fill=white}]
		\node[draw, circle, font=\large] (y) at  (-1, 1) {$Y$};
		\node[draw, circle, font=\large] (x1) at  (0,2) {$X_1$};
		\node[draw, circle, font=\large] (x2) at  (0,1) {$X_2$};
		\node[draw, circle, font=\large] (x3) at  (0,0) {$X_3$};
		\draw[->, black] (x1) -- (x2);
		\draw[->, black] (x2) -- (x3);
		\draw[->, black] (x3) -- (y);
		
		\node[draw, circle, font=\large] (yr) at  (3,1) {$\hat{Y}$};
		\node[draw, circle, font=\large] (x1r) at  (2,2) {$\underline{X}_1$};
		\node[draw, circle, font=\large] (x2r) at  (2,1) {$\underline{X}_2$};
		\node[draw, circle, font=\large] (x3r) at  (2,0) {$\underline{X}_3$};
		\draw[->, black] (x2r) -- (yr);
		\draw[->, black] (x3r) -- (yr);
		
		\draw[->, black, dotted] (x1) -- (x1r);
		\draw[->, black, dotted] (x2) -- (x2r);
		\draw[->, black, dotted] (x3) -- (x3r);
		
		\draw [draw=gray] (-1.5,-.5) rectangle (.5,2.5);
		\draw [draw=gray] (1.3,-.5) rectangle (3.5,2.5);
		\node[](model) at (2.5, -1) {model};
		\node[](data) at (-.5, -1) {data};
	\end{tikzpicture}
}
\hfill
\subfigure[]{
\centering
\includegraphics[width=0.55\linewidth]{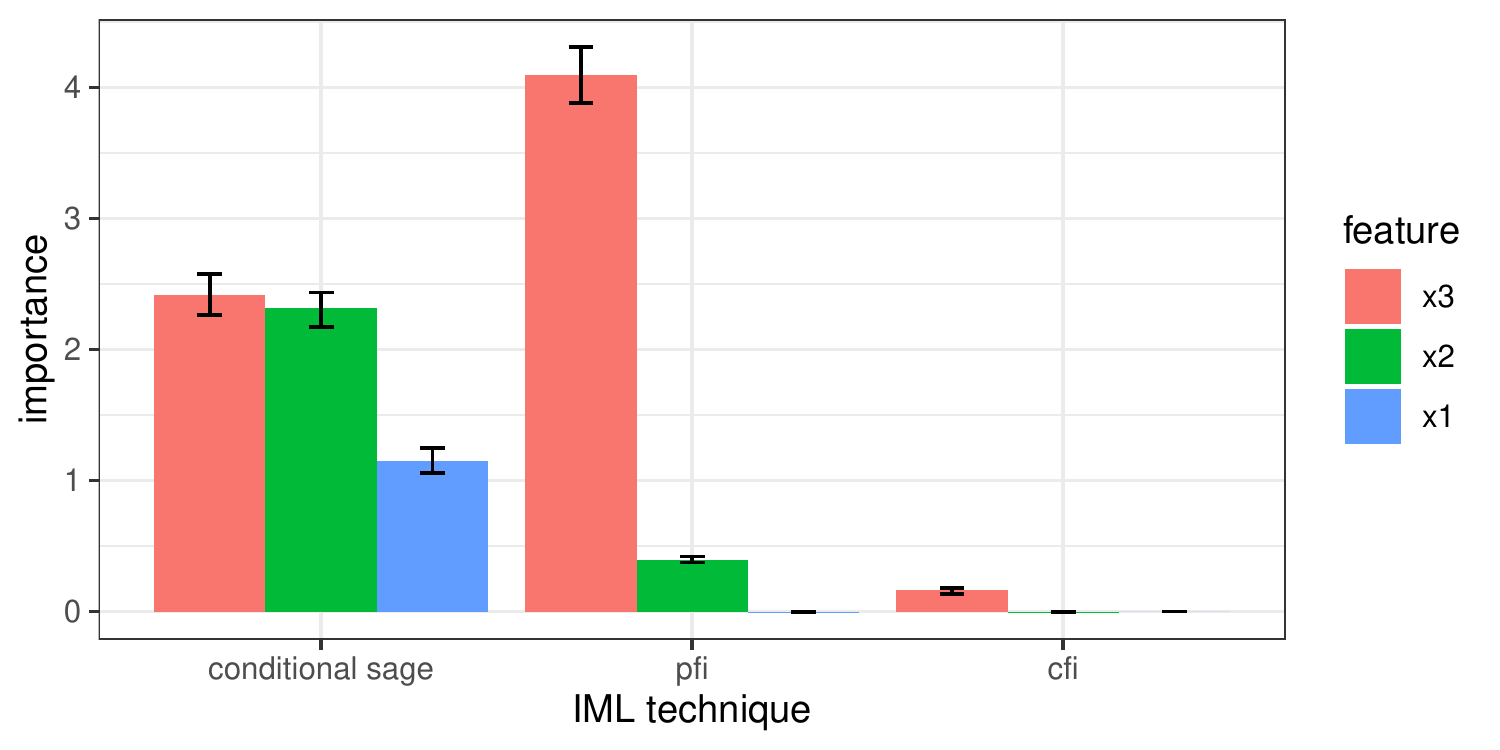}
}
    \caption{\textbf{Misunderstanding conditional interpretation}. A linear model was fitted on the data-generating process modeled using a linear Gaussian structural causal model. The entailed directed acyclic graph is depicted on the left. For illustrative purposes, the original model coefficients were updated such that not only feature $X_3$, but also feature $X_2$ is used by the model. PFI on test data considers both $X_3$ and $X_2$ to be relevant. In contrast, conditional feature importance variants either only consider $X_3$ to be relevant (CFI) or consider all features to be relevant (conditional SAGE value function).}
    \label{fig:cond-interpret}
\end{figure}
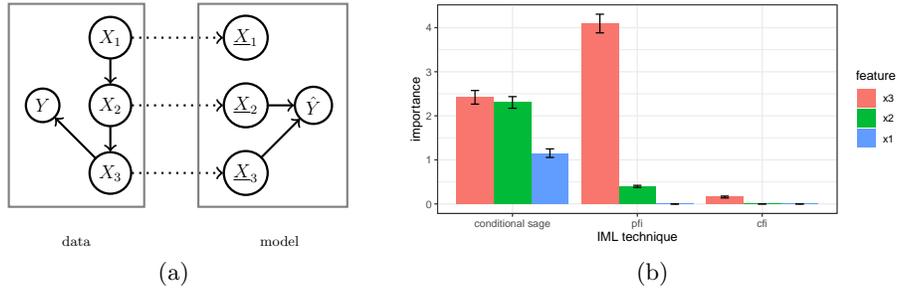

\subsubsection{Pitfall:} Conditional variants of interpretation techniques avoid extrapolation but require a different interpretation.
%
Interpretation methods that perturb features independently of others will extrapolate under dependent features but provide insight into the model's mechanism \cite{janzing2019feature,konig2021decomposition}. Therefore, these methods are said to be true to the model but not true to the data \cite{chen2020true}.

For feature effect methods such as the PDP, the plot can be interpreted as the isolated, average effect the feature has on the prediction.
For the PFI, the importance can be interpreted as the drop in performance when the feature's information is \enquote{destroyed} (by perturbing it). Marginal SHAP value functions \cite{lundberg2017unified} quantify a feature's contribution to a specific prediction, and marginal SAGE value functions \cite{NEURIPS2020_c7bf0b7c} quantify a feature's contribution to the overall prediction performance. All the aforementioned methods extrapolate under dependent features (see also Section~\ref{sec:sampling}), but satisfy sensitivity, i.e. are zero if a feature is not used by the model \cite{janzing2019feature,NEURIPS2020_c7bf0b7c,konig2021decomposition}.

Conditional variants of these interpretation methods do not replace feature values independently of other features, but in such a way that they conform to the conditional distribution.
This changes the interpretation as the effects of all dependent features become entangled. Depending on the method, conditional sampling leads to a more or less restrictive notion of relevance.

For example, for dependent features, the Conditional Feature Importance (CFI) \cite{candes2018panning,Watson2019,molnar2020model,strobl2008conditional} answers the question:
``How much does the model performance drop if we permute a feature, \textit{but given that we know the values of the other features?}" \cite{strobl2008conditional,konig2021relative,molnar2020model}. \footnote{While for CFI the conditional independence of the feature of interest $X_j$ with the target $Y$ given the remaining features $X_{-j}$ ($Y \perp X_j | X_{-j}$) is already a sufficient condition for zero importance, the corresponding PFI may still be nonzero \cite{konig2021relative}.} Two highly dependent features might be individually important (based on the unconditional PFI), but have a very low conditional importance score because the information of one feature is contained in the other and vice versa.

In contrast, the conditional variant of PDP, called marginal plot or M-plot \cite{apley2016visualizing}, violates sensitivity, i.e. may even show an effect for features that are not used by the model. This is because for M-plots, the feature of interest is not sampled conditionally on the remaining features, but rather the remaining features are sampled conditionally on the feature of interest. As a consequence, the distribution of dependent covariates varies with the value of the feature of interest.
Similarly, conditional SAGE and conditional SHAP value functions sample the remaining features conditional on the feature of interest and therefore violate sensitivity \cite{janzing2019feature,sundararajan2019many,NEURIPS2020_c7bf0b7c,konig2021decomposition}.

We demonstrate the difference between PFI, CFI, and conditional SAGE value functions on a simulated example (Figure \ref{fig:cond-interpret}) where the data-generating mechanism is known. While PFI only considers features to be relevant if they are actually used by the model, SAGE value functions may also consider a feature to be important that is not directly used by the model if it contains information that the model exploits. CFI only considers a feature to be relevant if it is both mechanistically used by the model and contributes unique information about $Y$.

%
%
\subsubsection{Solution:} 
When features are highly dependent and conditional effects and importance scores are used, the practitioner must be aware of the distinct interpretation. Recent work formalizes the implications of marginal and conditional interpretation techniques \cite{chen2020true,janzing2019feature,konig2021relative,konig2021decomposition,NEURIPS2020_c7bf0b7c}. While marginal methods provide insight into the model's mechanism but are not true to the data, their conditional variants are not true to the model but provide insight into the associations in the data.

If joint insight into model and data is required, designated methods must be used.
ALE plots \cite{apley2016visualizing} provide interval-wise unconditional interpretations that are true to the data. They have been criticized to produce non-intuitive results for certain data-generating mechanisms \cite{groemping2020model}. Molnar et al. \cite{molnar2020model} propose a subgroup-based conditional sampling technique that allows for group-wise marginal interpretations that are true to model and data and that can be applied to feature importance and feature effects methods such as conditional PDPs and CFI. For feature importance, the DEDACT framework \cite{konig2021decomposition} allows to decompose conditional importance measures such as SAGE value functions into their marginal contributions and vice versa, thereby allowing global insight into both: the sources of prediction-relevant information in the data as well as into the feature pathways by which the information enters the model.

\subsubsection{Open Issues:} The quality of conditional IML techniques depends on the goodness of the conditional sampler. Especially in continuous, high-dimensional settings, conditional sampling is challenging. More research on the robustness of interpretation techniques regarding the quality of the sample is required.
\section{Misleading Interpretations due to Feature Interactions}
\label{sec:interaction}
\subsection{Misleading Feature Effects due to Aggregation}
\subsubsection{Pitfall:}
Global interpretation methods, such as PDP or ALE plots, visualize the average effect of a feature on a model's prediction. However, they can produce misleading interpretations when features interact. 
Figure~\ref{fig:interact1} A and B show the marginal effect of features $X_1$ and $X_2$ of the below-stated simulation example. While the PDP of the non-interacting feature $X_1$ seems to capture the true underlying effect of $X_1$ on the target quite well (A), the global aggregated effect of the interacting feature $X_2$ (B) shows almost no influence on the target, although an effect is clearly there by construction.

%
\subsubsection{Solution:} 
\begin{figure}
    \centering
    \includegraphics[width=\textwidth]{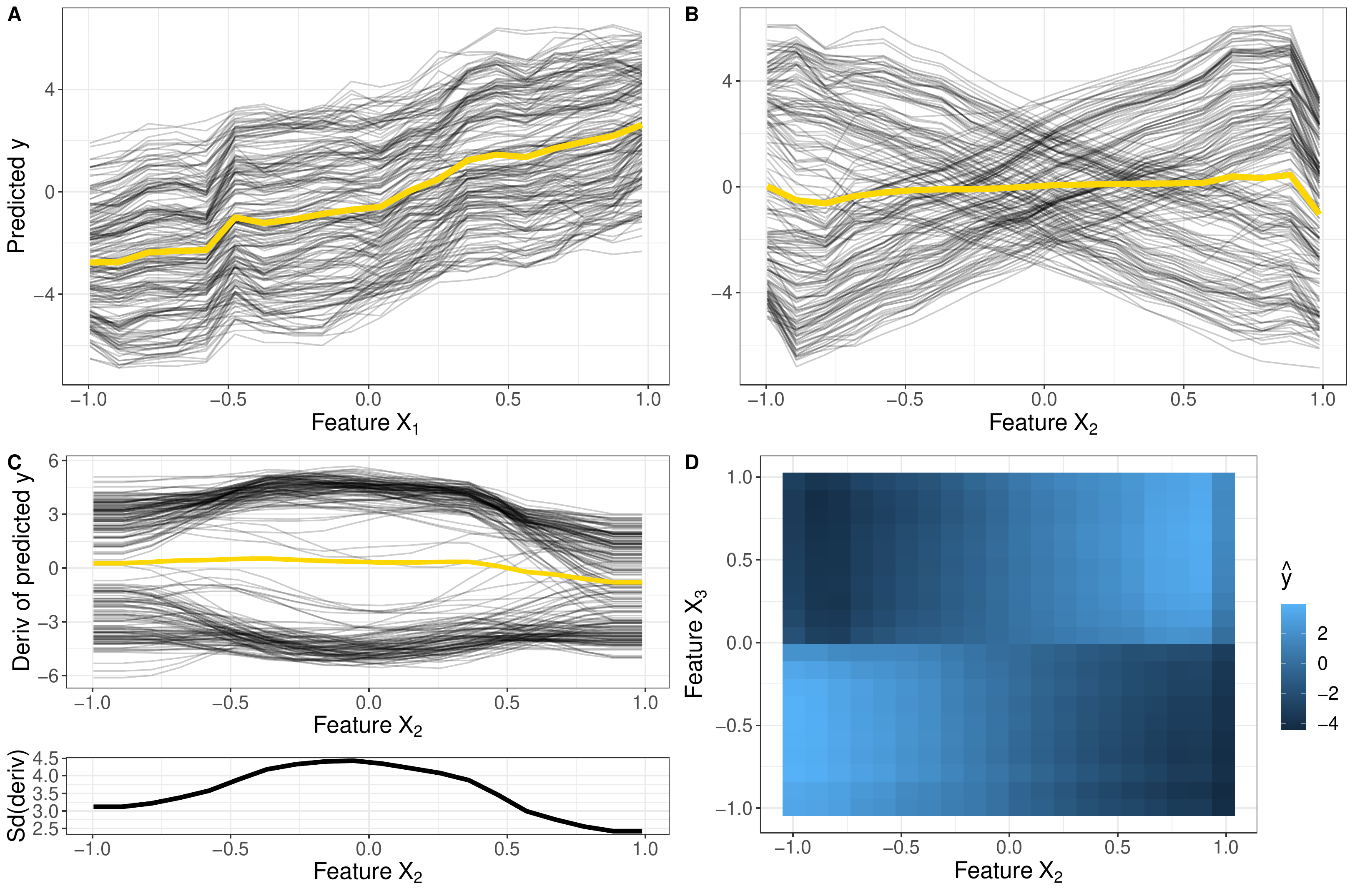}
    \caption{\textbf{Misleading effect due to interactions}. Simulation example with interactions: $Y= 3X_1 - 6X_2 + 12X_2 \mathbbm{1}_{(X_3 \geq 0)} + \epsilon$ with $X_1,X_2,X_3 \stackrel{i.i.d.}{\sim} U[-1,1]$ and $\epsilon \stackrel{i.i.d.}{\sim} N(0, 0.3)$. A random forest with 500 trees is fitted on 1000 observations. Effects are calculated on 200 randomly sampled (training) observations. \textbf{A, B:} PDP (yellow) and ICE curves of $X_1$ and $X_2$; \textbf{C:} Derivative ICE curves and their standard deviation of $X_2$; \textbf{D:} 2-dimensional PDP of $X_2$ and $X_3$.}
    \label{fig:interact1}
\end{figure}

For the PDP, we recommend to additionally consider the corresponding ICE curves \cite{goldstein2015peeking}.
While PDP and ALE average out interaction effects, ICE curves directly show the heterogeneity between individual predictions. 
Figure~\ref{fig:interact1} A illustrates that the individual marginal effect curves all follow an upward trend with only small variations. Hence, by aggregating these ICE curves to a global marginal effect curve such as the PDP, we do not lose much information. However, when the regarded feature interacts with other features, such as feature $X_2$ with feature $X_3$ in this example, then marginal effect curves of different observations might not show similar effects on the target. Hence, ICE curves become very heterogeneous, as shown in Figure~\ref{fig:interact1} B. In this case, the influence of feature $X_2$ is not well represented by the global average marginal effect.
Particularly for continuous interactions where ICE curves start at different intercepts, we recommend the use of derivative or centered ICE curves, which eliminate differences in intercepts and leave only differences due to interactions \cite{goldstein2015peeking}. 
Derivative ICE curves also point out the regions of highest interaction with other features. For example, Figure~\ref{fig:interact1} C indicates that predictions for $X_2$ taking values close to $0$ strongly depend on other features' values.
While these methods show that interactions are present with regards to the feature of interest but do not reveal other features with which it interacts, the 2-dimensional PDP or ALE plot are options to visualize 2-way interaction effects. The 2-dimensional PDP in Figure~\ref{fig:interact1} D shows that predictions with regards to feature $X_2$ highly depend on the feature values of feature $X_3$.
%
Other methods that aim to gain more insights into these visualizations are based on clustering homogeneous ICE curves, such as visual interaction effects (VINE) \cite{britton2019vine} or \cite{zhang2021interpreting}. As an example, in Figure~\ref{fig:interact1} B, it would be more meaningful to average over the upward and downward proceeding ICE curves separately and hence show that the average influence of feature $X_2$ on the target depends on an interacting feature (here: $X_3$). Work by Zon et al. \cite{ZonDIVP18} followed a similar idea by proposing an interactive visualization tool to group Shapley values with regards to interacting features that need to be defined by the user.

\subsubsection{Open Issues:} The introduced visualization methods are not able to illustrate the type of the underlying interaction and most of them are also not applicable to higher-order interactions.

\subsection{Failing to Separate Main from Interaction Effects}
\subsubsection{Pitfall:}
Many interpretation methods that quantify a feature's importance or effect cannot separate an interaction from main effects.
The PFI, for example, includes both the importance of a feature and the importance of all its interactions with other features \cite{casalicchio2019visualfi}. 
Also local explanation methods such as LIME and Shapley values only provide additive explanations without separation of main effects and interactions \cite{gosiewska2019not}.

\subsubsection{Solution:}
Functional ANOVA introduced by \cite{hooker2004VIN} is probably the most popular approach to decompose the joint distribution into main and interaction effects. 
Using the same idea, the H-Statistic \cite{friedman2008} quantifies the interaction strength between two features or between one feature and all others by decomposing the 2-dimensional PDP into its univariate components.
The H-Statistic is based on the fact that, in the case of non-interacting features, the 2-dimensional partial dependence function equals the sum of the two underlying univariate partial dependence functions.
Another similar interaction score based on partial dependencies is defined by \cite{greenwell2018}. Instead of decomposing the partial dependence function, \cite{Oh2019interaction} uses the predictive performance to measure interaction strength.
Based on Shapley values, Lundberg et al. \cite{lundberg2018consistent} proposed SHAP interaction values, and Casalicchio et al. \cite{casalicchio2019visualfi} proposed a fair attribution of the importance of interactions to the individual features.

Furthermore, Hooker \cite{hooker_generalized_FANOVA} considers dependent features and decomposes the predictions in main and interaction effects. A way to identify higher-order interactions is shown in \cite{hooker2004VIN}.
%
%
%

\subsubsection{Open Issues:} Most methods that quantify interactions are not able to identify higher-order interactions and interactions of dependent features.
Furthermore, the presented solutions usually lack automatic detection and ranking of all interactions of a model. Identifying a suitable shape or form of the modeled interaction is not straightforward as interactions can be very different and complex, e.g., they can be a simple product of features (multiplicative interaction) or can have a complex joint non-linear effect such as smooth spline surface. 
\section{Ignoring Model and Approximation Uncertainty}
\label{sec:uncertainty}
\subsubsection{Pitfall:} 
Many interpretation methods only provide a mean estimate but do not quantify uncertainty.
Both the model training and the computation of interpretation are subject to uncertainty.
The model is trained on (random) data, and therefore should be regarded as a random variable.
Interpretation methods are often defined in terms of expectations over the data (PFI, PDP, Shapley values, ...), but are approximated using Monte Carlo integration.
Ignoring these two sources of uncertainty can result in the interpretation of noise and non-robust results.
The true effect of a feature may be flat, but -- purely by chance, especially on smaller datasets -- the Shapley value might show an effect.
This effect could cancel out once averaged over multiple model fits.
\begin{figure}
    \centering
    \includegraphics[width=\textwidth]{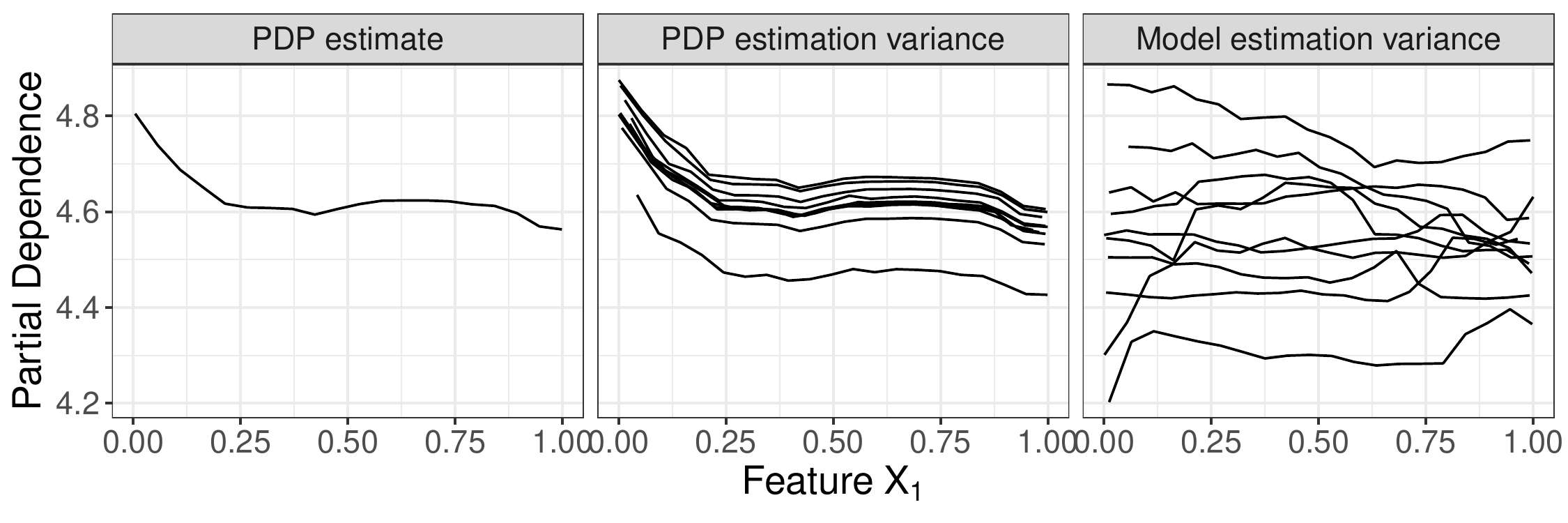}
    \caption{\textbf{Ignoring model and approximation uncertainty}. PDP for $X_1$ with $Y=0\cdot X_1 + \sum_{j=2}^{10} X_j + \epsilon_i$ with $X_1,\ldots,X_{10} \sim U[0,1]$ and $\epsilon_i \sim N(0, 0.9)$. \textbf{Left:} PDP for $X_1$ of a random forest trained on 100 data points. \textbf{Middle:} Multiple PDPs (10x) for the model from left plots, but with different samples (each n=100) for PDP estimation. \textbf{Right:} Repeated (10x) data samples of n=100 and newly fitted random forest.}
    \label{fig:variance}
\end{figure}
Figure~\ref{fig:variance} shows that a single PDP (first plot) can be misleading because it does not show the variance due to PDP estimation (second plot) and model fitting (third plot).
If we are not interested in learning about a specific model, but rather about the relationship between feature $X_1$ and the target (in this case), we should consider the model variance.

\subsubsection{Solution:} By repeatedly computing PDP and PFI with a given model, but with different permutations or bootstrap samples, the uncertainty of the estimate can be quantified, for example in the form of confidence intervals.
For PFI, frameworks for confidence intervals and hypothesis tests exist \cite{Watson2019,altmann2010permutation}, but they assume a fixed model.
If the practitioner wants to condition the analysis on the modeling process and capture the process' variance instead of conditioning on a fixed model, PDP and PFI should be computed on multiple model fits.

\subsubsection{Open Issues:} While Moosbauer et al. \cite{moosbauer2021xautoml} derived confidence bands for PDPs for probabilistic ML models that cover the model's uncertainty, a general model-agnostic uncertainty measure for feature effect methods such as ALE \cite{apley2016visualizing} and PDP \cite{friedman1991multivariate} has (to the best of our knowledge) not been introduced yet. 
\section{Failure to Scale to High-Dimensional Settings}
\label{sec:multiple}

\subsection{Human-Intelligibility of High-Dimensional IML Output}
\label{sec:high-dimensional}
\subsubsection{Pitfall:}
Applying IML methods naively to high-dimensional 
datasets (e.g. visualizing feature effects or computing importance scores on feature level) leads to an overwhelming and high-dimensional IML output, which impedes human analysis.
Especially interpretation methods that are based on visualizations make it difficult for practitioners in high-dimensional settings to focus on the most important insights.



\subsubsection{Solution:}
A natural approach is to reduce the dimensionality before applying any IML methods.
Whether this facilitates understanding or not depends on the possible semantic interpretability of the resulting, reduced feature space -- as features can either be selected or dimensionality can be reduced by linear or non-linear transformations. 
Assuming that users would like to interpret in the original feature space, many feature selection techniques can be used \cite{guyon2003introduction}, resulting in much
sparser and consequently easier to interpret models.
Wrapper selection approaches are model-agnostic and algorithms like greedy forward selection or subset selection procedures \cite{kohavi1997wrappers,au2021grouped},
which start from an empty model and iteratively add relevant (subsets of) features if needed, even allow to measure the relevance of features for predictive performance. 
An alternative is to directly use models that implicitly perform feature selection such as LASSO \cite{tibshirani1996regression} or component-wise boosting \cite{schmid2008boosting} as they can produce sparse models with fewer features. 

When features can be meaningfully grouped in a data-driven or knowledge-driven way \cite{he2010}, applying IML methods directly to grouped features instead of single features is usually more time-efficient to compute and often leads to more appropriate interpretations.
Examples where features can naturally be grouped include the grouping of sensor data \cite{Chakraborty2008}, time-lagged features \cite{Lozano2009}, or one-hot-encoded categorical features and interaction terms \cite{Gregorutti2015}. Before a model is fitted, groupings could already be exploited for dimensionality reduction, for example by selecting groups of features by the group LASSO \cite{Yuan2006}. 

For model interpretation, various papers extended feature importance methods from single features to groups of features \cite{au2021grouped,Gregorutti2015,valentin2020interpreting,williamson2020unified}. In the case of grouped PFI, this means that we perturb the entire group of features at once and measure the performance drop compared to the unperturbed dataset. 
Compared to standard PFI, the grouped PFI does not break the association to the other features of the group, but to features of other groups and the target. This is especially useful when features within the same group are highly correlated (e.g. time-lagged features), but between-group dependencies are rather low. Hence, this might also be a possible solution for the extrapolation pitfall described in Section \ref{sec:sampling}.

We consider the PhoneStudy in \cite{Stachl2020} as an illustration. 
The PhoneStudy dataset contains 1821 features to analyze the link between human behavior based on smartphone data and participants' personalities. 
Interpreting the results in this use case seems to be challenging since features were dependent and single feature effects were either small or non-linear \cite{Stachl2020}. The features have been grouped in behavior-specific categories such as app-usage, music consumption, or overall phone usage.
Au et al. \cite{au2021grouped} calculated various grouped importance scores on the feature groups to measure their influence on a specific personality trait (e.g. conscientiousness).
Furthermore, the authors applied a greedy forward subset selection procedure via repeated subsampling on the feature groups and showed that combining app-usage features and overall phone usage features were most of the times sufficient for the given prediction task.

\subsubsection{Open Issues:}
The quality of a grouping-based interpretation strongly depends on the human intelligibility and meaningfulness of the grouping. If the grouping structure is not naturally given, then data-driven methods can be used. However, if feature groups are not meaningful (e.g. if they cannot be described by a super-feature such as app-usage), then subsequent interpretations of these groups are purposeless. One solution could be to combine feature selection strategies with interpretation methods. However, this remains an open issue that requires further research.

Existing research on grouped interpretation methods mainly focused on quantifying grouped feature importance, but the question of ``how a group of features influences a model's prediction" remains almost unanswered. Only recently, \cite{au2021grouped,brenning2021transforming,seedorff2021totalvis} attempted to answer this question by using dimension-reduction techniques (such as PCA) before applying the interpretation method. However, this is also a matter of further research.

\subsection{Computational Effort}

\subsubsection{Pitfall:}
Some interpretation methods do not scale linearly with the number of features.
For example, for the computation of exact Shapley values the number of possible coalitions \cite{NEURIPS2020_c7bf0b7c,lundberg2017unified}, or for a (full) functional ANOVA decomposition the number of components (main effects plus all interactions) scales with $\mathcal{O}(2^{p})$ \cite{hooker_generalized_FANOVA}.\footnote{Similar to the PDP or ALE plots, the functional ANOVA components describe individual feature effects and interactions.}


\subsubsection{Solution:}
For the functional ANOVA, a common solution is to keep the analysis to the main effects and selected 2-way interactions (similar for PDP and ALE). Interesting 2-way interactions can be selected by another method such as the H-statistic \cite{friedman2008}.
However, the selection of 2-way interactions requires additional computational effort.
Interaction strength usually decreases quickly with increasing interaction size, and one should only consider $d$-way interactions when all their $(d-1)$-way interactions were significant \cite{hooker2004VIN}.
For Shapley-based methods, an efficient approximation exists that is based on randomly sampling and evaluating feature orderings until the estimates converge. The variance of the estimates reduces in $\mathcal{O}(\frac{1}{m})$, where $m$ is the number of evaluated orderings \cite{NEURIPS2020_c7bf0b7c,lundberg2017unified}.

\subsection{Ignoring Multiple Comparison Problem}
\subsubsection{Pitfall:}
Simultaneously testing the importance of multiple features will result in false-positive interpretations if the multiple comparisons problem (MCP) is ignored.
The MCP is well known in significance tests for linear models and exists similarly in testing for feature importance in ML.
For example, suppose we simultaneously test the importance of 50 features (with the $H_0$-hypothesis of zero importance) at the significance level $\alpha = 0.05$. Even if all features are unimportant, the probability of observing that at least one feature is significantly important is
        $1 - \mathbb{P}(\text{`no feature important'}) = 1-(1-0.05)^{50} \approx 0.923$.
Multiple comparisons become even more problematic the higher the dimension of the dataset.

\subsubsection{Solution:} Methods such as Model-X knockoffs \cite{candes2018panning} directly control for the false discovery rate (FDR).
For all other methods that provide p-values or confidence intervals, such as PIMP (Permutation IMPortance) \cite{altmann2010permutation}, which is a testing approach for PFI, MCP is often ignored in practice to the best of our knowledge, with some exceptions\cite{stachl2019behavioral,Watson2019}.
One of the most popular MCP adjustment methods is the Bonferroni correction \cite{dunn1961correction}, which rejects a null hypothesis if its p-value is smaller than $\alpha / p$, with $p$ as the number of tests. 
It has the disadvantage that it increases the probability of false negatives \cite{perneger98correction}.
Since MCP is well known in statistics, we refer the practitioner to \cite{dickhaus2014correction} for an overview and discussion of alternative adjustment methods, such as the Bonferroni-Holm method \cite{holm1979correction}.

\begin{figure}
    \centering
    \includegraphics[width=1\textwidth]{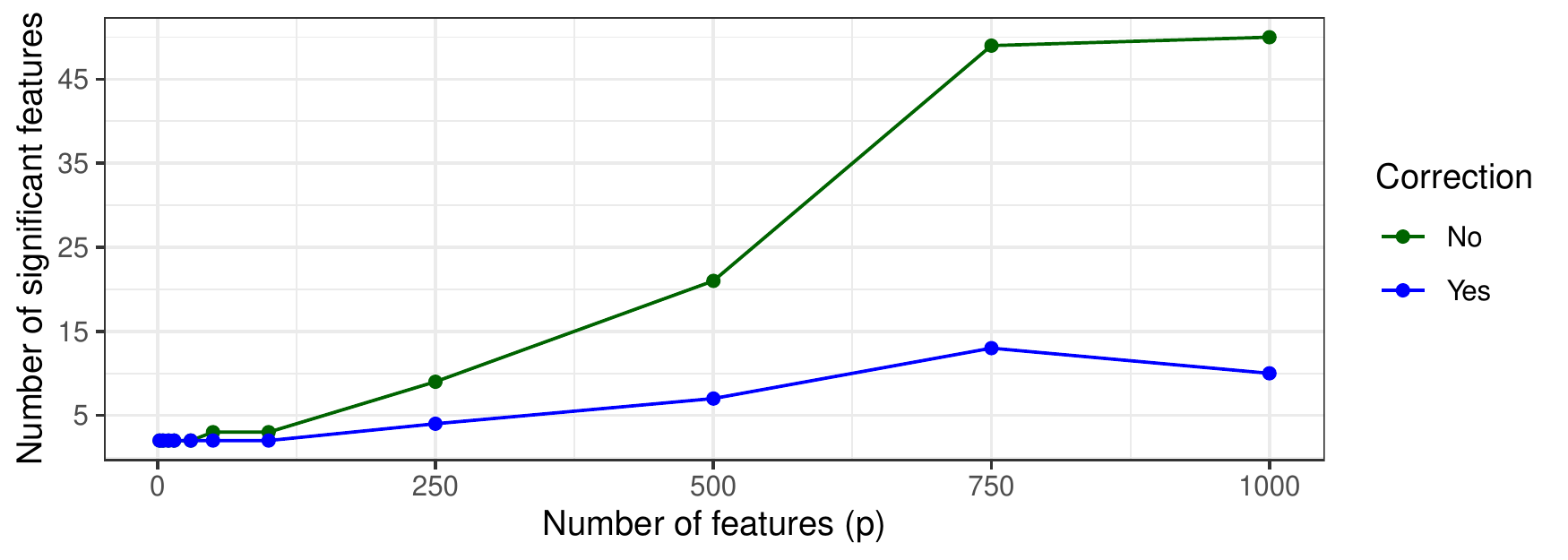}
    \caption{\textbf{Failure to scale to high-dimensional settings}. Comparison of the number of features with significant importance - once with and once without Bonferroni-corrected significance levels for a varying number of added noise variables. Datasets were sampled from $Y = 2X_1 + 2X_2^2 + \epsilon$ with $X_1, X_2,\epsilon \sim N(0, 1)$. $X_3, X_4, ..., X_p \sim N(0, 1)$ are additional noise variables with $p$ ranging between 2 and 1000. For each $p$, we sampled two datasets from this data-generating process -- one to train a random forest with 500 trees on and one to test whether feature importances differed from 0 using PIMP.  In all experiments, $X_1$ and $X_2$ were correctly identified as important.}
    \label{fig:bonferroni}
\end{figure}

As an example, in Figure~\ref{fig:bonferroni} we compare the number of features with significant importance measured by PIMP once with and once without Bonferroni-adjusted significance levels ($\alpha = 0.05$ vs. $\alpha = 0.05/p$). Without correcting for multi-comparisons, the number of features mistakenly evaluated as important grows considerably with increasing dimension, whereas Bonferroni correction results in only a modest increase.

\section{Unjustified Causal Interpretation}
\label{sec:causality}
\subsubsection{Pitfall:}
Practitioners are often interested in causal insights into the underlying data-generating mechanisms, which IML methods do not generally provide. Common causal questions include the identification of causes and effects, predicting the effects of interventions, and answering counterfactual questions \cite{pearl2018ladder}. For example, a medical researcher might want to identify risk factors or predict average and individual treatment effects \cite{Koenig2019}. In search of answers, a researcher can therefore be tempted to interpret the result of IML methods from a causal perspective.

However, a causal interpretation of predictive models is often not possible. Standard supervised ML models are not designed to model causal relationships but to merely exploit associations.
A model may therefore rely on causes and effects of the target variable as well as on variables that help to reconstruct unobserved influences on $Y$, e.g. causes of effects \cite{Weichwald2015}. Consequently, the question of whether a variable is relevant to a predictive model (indicated e.g. by PFI $>$ 0) does not directly indicate whether a variable is a cause, an effect, or does not stand in any causal relation to the target variable.
Furthermore, even if a model would rely solely on direct causes for the prediction, the causal structure between features must be taken into account. Intervening on a variable in the real world may affect not only $Y$ but also other variables in the feature set. Without assumptions about the underlying causal structure, IML methods cannot account for these adaptions and guide action \cite{Karimi2020}.

\begin{figure}[H]
	\centering
	\begin{tikzpicture}[thick, scale=1, every node/.style={scale=0.6, line width=0.3mm, black, fill=white}]
		\node[draw, circle, font=\large] (y) at  (0,0) {$Y$};
		\node[draw, circle, font=\large] (x1) at  (-1,.7) {$X_1$};
		\node[draw, circle, font=\large] (x2) at  (1,.7) {$X_2$};
		\node[draw, circle, font=\large] (x3) at  (-2,0) {$X_3$};
		\node[draw, circle, font=\large] (x4) at  (-1,-.7) {$X_4$};
		\node[draw, circle, font=\large] (x5) at  (1,-.7) {$X_5$};
		\draw[->, black] (x1) -- (y);
		\draw[->, black] (x1) -- (x2);
		\draw[->, black] (x2) -- (y);
		\draw[->, black] (y) -- (x4);
		\draw[->, black] (y) -- (x5);
		\draw[->, black] (x3) -- (x4);
	\end{tikzpicture}
	\caption{Causal graph}
\label{fig:scm-data-model}
\end{figure}
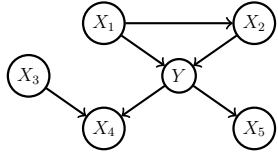

As an example, we constructed a dataset by sampling from a structural causal model (SCM), for which the corresponding causal graph is depicted in Figure \ref{fig:scm-data-model}. All relationships are linear Gaussian with variance $1$ and coefficients $1$. For a linear model fitted on the dataset, all features were considered to be relevant based on the model coefficients ($\hat{y} = 0.329 x_1 + 0.323 x_2 - 0.327 x_3 +  0.342 x_4 + 0.334 x_5$, $R^2=0.943$), although $x_3$, $x_4$ and $x_5$ do not cause $Y$.

\subsubsection{Solution:} The practitioner must carefully assess whether sufficient assumptions can be made about the underlying data-generating process, the learned model, and the interpretation technique. If these assumptions are met, a causal interpretation may be possible.
The PDP between a feature and the target can be interpreted as the respective average causal effect if the model performs well and the set of remaining variables is a valid adjustment set \cite{zhao2019causal}.
When it is known whether a model is deployed in a causal or anti-causal setting -- i.e. whether the model attempts to predict an effect from its causes or the other way round -- a partial identification of the causal roles based on feature relevance is possible (under strong and non-testable assumptions) \cite{Weichwald2015}. 
Designated tools and approaches are available for causal discovery and inference \cite{Peters2017book}.

\subsubsection{Open Issues:} The challenge of causal discovery and inference remains an open key issue in the field of ML. Careful research is required to make explicit under which assumptions what insight about the underlying data-generating mechanism can be gained by interpreting an ML model.

\section{Discussion}
\label{sec:discussion}

In this paper, we have reviewed numerous pitfalls of local and global model-agnostic interpretation techniques, e.g. in the case of bad model generalization, dependent features, interactions between features, or causal interpretations.
Although this exploration of pitfalls is far from complete, we believe that we cover common ones that pose a particularly high risk.
We hope to encourage a more cautious approach when interpreting ML models in practice, to point practitioners to already (partially) available solutions, and to stimulate further research on these issues.
The stakes are high: ML algorithms are increasingly used for socially relevant decisions, and model interpretations play an important role in every empirical science.
Therefore, we believe that users can benefit from concrete guidance on properties, dangers, and problems of IML techniques -- especially as the field is advancing at high speed.
We need to strive towards a recommended, well-understood set of tools, which will in turn require much more careful research.
This especially concerns the meta-issues of comparisons of IML techniques, IML diagnostic tools to warn against misleading interpretations, and tools for analyzing multiple dependent or interacting features.

\bibliography{paper}
\bibliographystyle{splncs04}

\end{document}